\documentclass{article}

\usepackage{float}
\usepackage{threeparttable}
\usepackage{longtable}
\usepackage{graphicx}
\usepackage{multirow}
\usepackage[utf8]{inputenc}
\usepackage{titlesec} 
\usepackage{changepage} 
\usepackage{makecell}
\usepackage{authblk} 
\usepackage{arxiv}
\usepackage{amsthm}
\usepackage[utf8]{inputenc} 
\usepackage[T1]{fontenc}    

\usepackage[hidelinks]{hyperref}

\usepackage{url}            
\usepackage{booktabs}       
\usepackage{amsfonts}       
\usepackage{nicefrac}       
\usepackage{microtype}      
\usepackage{lipsum}
\usepackage{graphicx}
\graphicspath{ {./images/} }

\usepackage{parskip}
\usepackage{graphicx}
\usepackage{adjustbox}

\usepackage{amsmath}
\usepackage{titlesec}
\title{Can Frontier LLMs Replace Annotators in Biomedical Text Mining? Analyzing Challenges and Exploring Solutions}

\author[1]{Yichong Zhao\thanks{Email: eric-yc-zhao@g.ecc.u-tokyo.ac.jp}\,\,\,}
\author[1,2]{Susumu Goto\thanks{Corresponding author: goto@dbcls.rois.ac.jp}\,\,\,}

\affil[1]{Department of Computational Biology and Medical Sciences, Graduate School of Frontier Sciences, The University of Tokyo.}
\affil[2]{Database Center for Life Science, Joint Support-Center for Data Science Research, ROIS}

\begin{document}


\newpage

\maketitle
\begin{abstract}
Multiple previous studies have reported suboptimal performance of LLMs in biomedical text mining. By analyzing failure patterns in these evaluations, we identified three primary challenges for LLMs in biomedical corpora: (1) LLMs fail to learn implicit dataset-specific nuances from supervised data, (2) The common formatting requirements of discriminative tasks limit the reasoning capabilities of LLMs particularly for LLMs that lack test-time compute, and (3) LLMs struggle to adhere to annotation guidelines and match exact schemas, which hinders their ability to understand detailed annotation requirements which is essential in biomedical annotation workflow. We experimented with prompt engineering techniques targeted to the above issues, and developed a pipeline that dynamically extracts instructions from annotation guidelines. Our results show that frontier LLMs can approach or surpass the performance of state-of-the-art (SOTA) BERT-based models with minimal reliance on manually annotated data and without fine-tuning. Furthermore, we performed model distillation on a closed-source LLM, demonstrating that a BERT model trained exclusively on synthetic data annotated by LLMs can also achieve a practical performance. Based on these findings, we explored the feasibility of partially replacing manual annotation with LLMs in production scenarios for biomedical text mining.
\end{abstract}

\section{Introduction}\label{sec:intro}
\subsection{Background: Biomedical Text Mining and the Role of Language Models}\label{subsec:1.1}
Biomedical literature contains valuable knowledge essential for building curated biological databases and enabling data-driven discovery. Information extraction from this literature has traditionally relied on manual annotation, where domain experts develop annotation guidelines, and human annotators label data accordingly. This labor-intensive process remains indispensable in high-precision scenarios, such as maintaining resources like UniProtKB/Swiss-Prot and the Gene Ontology databases\cite{consortium2023gene}.

\begin{figure}[H]
\centering
\hspace*{-20px}
\includegraphics[width=1.15\textwidth]{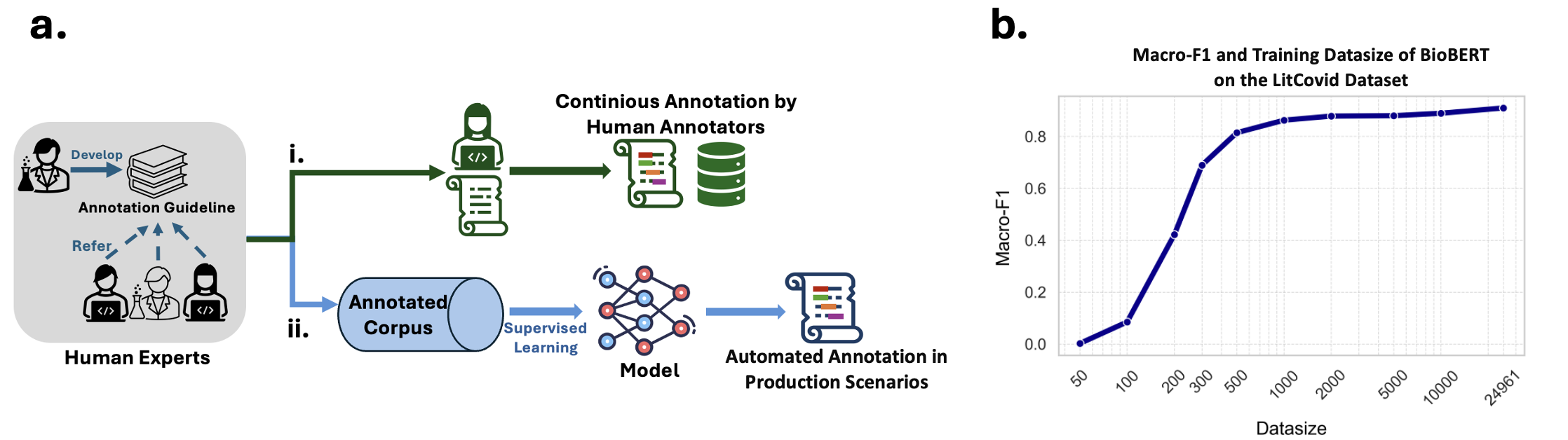}
\caption{\textbf{The Reliance of Biomedical Text Mining Tasks on Human Annotators}}
\label{fig:1} 
\raggedright\textbf {a.}  This panel illustrates two paradigms of biomedical text mining in real-world applications: (i) Continuous annotation by human experts, commonly seen in scenarios requiring high annotation precision, such as database curation. (ii) The constructions of supervised datasets by human experts, which are then used to train machine learning models for annotating large-scale or continuously produced texts. \\\textbf{b.} This panel demonstrates the dependency of machine learning performance on the amount of training data in the multi-label text classification dataset LitCovid. Achieving satisfactory model performance requires the manual annotation of hundreds or even thousands of data samples.
\end{figure}

The labor-intensive nature of manual annotation has driven the adoption of automated biomedical text mining approaches—such as named entity recognition (NER), relation extraction (RE), and multi-label classification\cite{zhao2021recent}. Tools like PubTator\cite{ wei2024pubtator} and LitCovid\cite{chen2023litcovid} demonstrate the feasibility of applying machine learning to assist with information extraction and document triaging. Complementing this trend is the rise of BERT-based models, which have become the mainstream approach in recent years\cite{devlin2018bert}. BERT usually undergone self-supervised pretraining and are then fine-tuned on task-specific supervised datasets to perform diverse biomedical text mining tasks. However, these methods remain fundamentally dependent on manually annotated data (Figure~\ref{fig:1}).


The emergence of large language models (LLMs) like ChatGPT offers a potential paradigm shift. LLMs are typically based on Transformer decoder architecture and are pre-trained on large-scale copora using next-token prediction to develop text generation capabilities\cite{brown2020language}. Additionally, LLMs undergo alignment training to improve their ability to follow human instructions, engage in logical reasoning, and adapt to diverse tasks without specialized training.

As early as 2022, studies began applying models like GPT-3 to information extraction, showing promising results in clinical domain\cite{agrawal2022large}. Following the release of ChatGPT, Qijie Chen et al. benchmarked GPT-3.5 on various tasks in the biomedical literature\cite{chen2023extensive}. While the model outperformed baselines in {Q\&A} tasks, its performance on information extraction tasks lagged significantly behind those of fine-tuned models. Similarly, a comprehensive evaluation of LLMs across 26 datasets conducted by Jahan et al. confirmed that these models struggle to match the performance of fine-tuned BERTs in discriminative tasks\cite{jahan2024comprehensive}. Further studies, such as those by Qingyu Chen et al. highlighted that even advanced proprietary models like GPT-4 which excel in biomedical generative tasks, exhibit notable performance deficits compared to fine-tuned BERT models in discriminative tasks\cite{chen2023large}. In some datasets, GPT-4 achieved only 40\% of the SOTA performance. They furthermore analyzed the errors made by LLMs, but those analysis primarily focused on output format issues, without delving into the reasoning processes of the models. Overall, those studies revised that despite advancements in LLMs, fine-tuned BERT models remain superior for biomedical text mining tasks, highlighting a gap that still needs to be addressed.

\subsection{Contribution of Our Work}
We target three core biomedical text mining tasks: named entity recognition (NER), relation extraction (RE), and multi-label text classification. NER extracts biomedical entities (e.g., genes, drugs, diseases) from text. RE identifies relationships among these entities, such as drug–disease interactions. Multi-label classification assigns multiple relevant labels to biomedical texts at the sentence or document level.

We first reproduced the previous evaluations of LLM in biomedical domain to analyze specific failure patterns, then introduced series of prompt engineering methods and a information retrieval pipeline specially targeted those issues we identified:
 
(1) We revealed a significant discrepancy between the distribution of LLM outputs and the ground truth label across various dimensions such as the number and distribution of output labels, average span length of entity mentions. 
We assume this difference arises because supervised training implicitly captures dataset-specific features. To address this, we introduced an automatic prompt optimization technique targeted to find a dataset-specific prompt narrowing the gap between LLM-predicted labels and the correct labels.

(2) Structured batch outputs are often favored for their convenience in parsing and downstream analysis in biomedical text mining tasks. However, this approach may unintentionally limit the ability of LLMs to showcase their reasoning processes during inference, which, as observed in our analysis and other studies, can negatively affect their overall performance\cite{tam2024let, strutureblog1}. To mitigate that, we developed a two-step inference approach targeted to preserve chain-of-thought reasoning during structured outputs\cite{wei2022chain}. 
Additionally, in the later stages of our study, several commercial models emerged that support test-time computation, enabling automated reasoning before the final output. We also conducted evaluations on one of the representative reasoning models.

(3) Many LLM failures were traced to insufficient understanding of task-specific annotation guidelines, which are often several pages to tens of pages long, specify crucial aspects such as task requirements, annotation categories, hierarchies, granularities, and instructions for handling edge cases. These details that are critical for accurate annotation but were largely overlooked in prior evaluations of LLMs. 
However, since annotation guidelines are often lengthy, LLMs struggle to identify the most relevant sections when these guidelines are directly appended to the prompt, resulting in suboptimal performance. To address this, we manually divided the annotation guidelines into smaller text chunks and developed a retrieval-augmented pipeline to dynamically retrieve the most relevant sections based on task context.

Utilizing the above approach, we achieved significant improvements in biomedical text mining of LLM performance, surpassing the SOTA on 2 datasets and demonstrating practical performance levels on the remaining.

Though some studies have shown that fine-tuning (including instruction tuning) may also be an effective approach\cite{keloth2024advancing, zhang2024fine}, the primary reason we did not adopt a fine-tuning approach is the substantial computational resources required for fine-tuning and deploying a large language model. Moreover, it is widely recognized that advanced prompt engineering can substantially improve the performance of LLMs, enabling general-domain models to achieve or even surpass domain-specific models\cite{nori2023can}.



Additionally, deploying LLMs is highly challenging, and relying on cloud services introduces significant financial costs and potential data security risks. To address this, we adopted a workflow inspired by annotation pipelines in current production scenarios: human-annotated data is used to train machine learning models, which then continuously perform inference. We treated the LLM with its output parser as a discriminative model and distilled it into a lighter BERT-based model by weak supervision learning on two datasets. We collected literature following the same procedure as in the original studies and used the LLM to generate annotated data, which were used to fine-tune a BERT model. The distilled BERT model achieved performance close to the previous SOTA. 

Our results suggest that for emerging text mining demands, relying solely on LLM inference in production environments or using an LLM+BERT distillation pipeline may be a practicall replacement of the widely adopted workflow of human annotation followed by machine learning. 

\section{Problem Analysis: LLM in Biomedical Text Mining}

\subsection{Biomedical Tasks and Datasets}

 
Table 1 summarizes the datasets used in our experiments along with their various characteristics. For our baselines, we selected well-known SOTA foundation models such as BioBERT\cite{lee2020biobert} and BioLinkBERT\cite{yasunaga2022linkbert}, which have demonstrated strong performance across a wide range of tasks. However, since we were unable to identify any prior study about BC5CDR-RE dataset utilizing BERT-based models, we conducted our own fine-tuning and evaluation of BioBERT on this dataset in addition to reporting the results of the current SOTA result which was achieved by a fine-tuned BioGPT\cite{luo2022biogpt}. 

\begin{table}[]
\caption{Datasets and Their Key Features}
\label{tab:datasum}
\renewcommand{\arraystretch}{1.2}
\hspace*{-2cm}
\begin{tabular}{llllll}
\hline
Tasks                                                                                  & Dataset                                                                   & \begin{tabular}[c]{@{}l@{}}Datasize \& Split\\ (Train/Valid/Test)\end{tabular}                                & \begin{tabular}[c]{@{}l@{}}Evaluation\\ Metrics \end{tabular}         & Baselines         &\begin{tabular}[c]{@{}l@{}}Annotation\\ Guideline \end{tabular} \\ \hline
\multirow{2}{*}{\begin{tabular}[c]{@{}l@{}}Named Entity\\ Recoginization\end{tabular}} & \begin{tabular}[c]{@{}l@{}}BC5CDR-Chemical/Disease\cite{li2016biocreative}\end{tabular} & 4560/4581/4797                                                                                                & \multirow{2}{*}{Entity-Level} & BioLinkBERT-Large & Y (8 Pages)  \\ & NCBI-Disease\cite{dougan2014ncbi} & 5424/923/940  &   & BioBERT-Large v1.2  & Y (5 Pages)          \\ 
\hline
\multirow{2}{*}{Relation Extraction}                                                   & BioRED\cite{luo2022biored}                                                                    & 4873/1162/1163                                                                                                & \multirow{2}{*}{Macro-Level}  & BiomedBERT $^*$         & Y (16 Pages)         \\
                                                                                       & BC5CDR-RE\cite{li2016biocreative}                                                                 & 1323/1343/1375                                                                                                   &                               & BioLinkBERT-Large, BioGPT & Y (8 Pages)          \\ \hline
\multirow{2}{*}{\begin{tabular}[c]{@{}l@{}}Multi-label \\ Classification\end{tabular}} & LitCovid                                                                  & 24960/6239/2500$^*$$^*$& \multirow{2}{*}{Macro-Level}  & BioBERT-Large v1.2           & N                    \\
                                                                                       & HoC                                                                       & 1108/157/315                                                                                                  &                               & BioLinkBERT-Large & N                    \\ \hline
\end{tabular}
\begin{tablenotes}
\item[$^*$] $^*$ \textit{When fine-tuning the baseline model BioBERT on the BC5CDR (RE) dataset, our initial exploration yielded catastrophic results as the model overwhelmingly predicted negative labels. To address this, we randomly sampled and reduced the number of negative labels in the training set. We achieved the results presented in the table when the ratio of negative to positive labels was reduced to 2:1, }
\item[$^*$] $^*$$^*$ \textit{Randomly sampled 500 test samples for LLM evaluation due to API cost constraints.}
\end{tablenotes}
\end{table}

\subsection{Challenges Faced by LLM}\label{sec:LLMchallenge}

\begin{figure*}[!t]
\centering
\hspace*{-20px}
\includegraphics[width=1\textwidth]{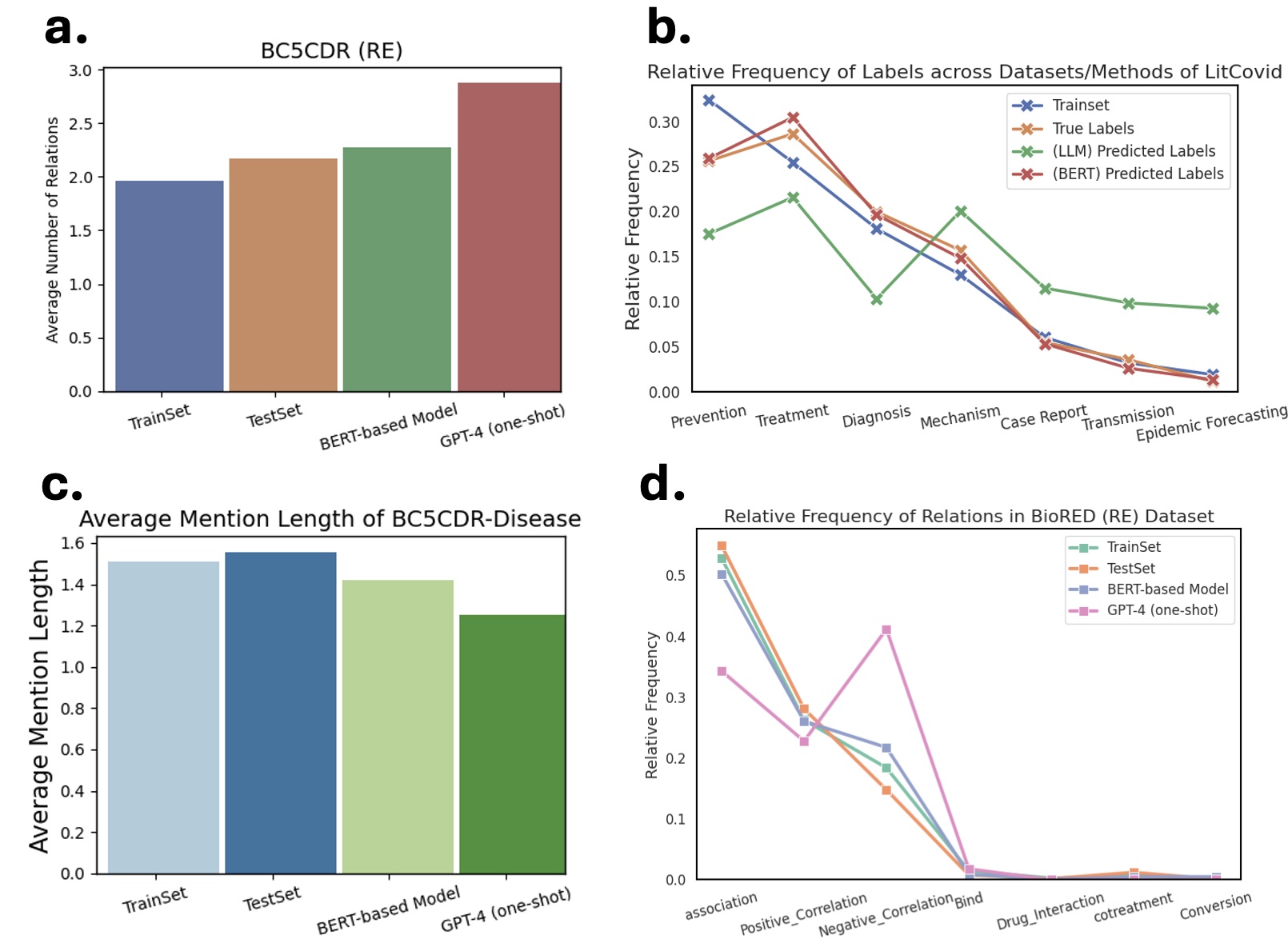}
\caption{\raggedright
\textbf{Models fine-tuned via supervised learning exhibit label distributions that align more closely with the training set and ground truth.}}
\label{fig:2}
\raggedright\textbf{a.} In the BC5CDR (RE) dataset, the training and test sets have an average of 1.96 and 2.16 CDI relationships per test instance, respectively. The predictions of a supervised BERT model average 2.27 CDI relationships per test instance, aligning more closely with the dataset’s characteristics. In contrast, GPT-4 predicts an average of 2.88 relationships, despite achieving higher performance (Macro-F1: 0.69 vs. 0.44).\\
\textbf{b.} Label frequency distribution in the LitCovid multi-label text classification dataset.\\
\textbf{c.} Average word count per entity mention in the BC5CDR-Disease named entity recognition dataset. GPT-4 outputs tend to select fewer words per entity, reflecting possible ambiguity in defining annotation boundaries.\\
\textbf{d.} Frequency distribution of relationship types in the BioRED relation extraction dataset.
\end{figure*}

In prior studies evaluating LLMs for biomedical text mining, the prompts typically included brief descriptions of the task type, labels, and output format\cite{chen2023large, chen2023extensive, jahan2024comprehensive}. We reproduced their approaches using GPT-4 (API version: 0613\cite{achiam2023gpt}), following Qingyu Chen's research) under a random 3-shot setting, and analyzed failure patterns of the outputs (Table 2). Our evaluation scores are comparable to or slightly higher than previous studies, which may be attributed to subtle differences in prompt text and improved robustness of the output parsers. However, LLM still significantly underperforms BERT-based models across all datasets except BC5CDR-RE.

We analyzed the following statistics for each tasks, to investigate the characteristics of outputs from LLMs:
\begin{itemize}
    \item Named entity recolonization: the number of entities per test instance and the average word length of each entity across the whole test set.
    \item Relation extraction: the number of relationships per test instance and the distribution of relationship categories.
    \item Multi-label text classification: the number of labels per test instance and the distribution of each label.
\end{itemize}
We selected four representative results shown in Figure \ref{fig:2}, with additional details provided in Figures S1 and S2. Across various metrics, BERT-based models generally produced outputs closer to the ground truth labels compared to GPT-4. Notably, GPT-4 achieved performance comparable to the SOTA model (fine-tuned BioGPT) and outperformed fine-tuned BioBERT on the BC5CDR-RE dataset. However, the distribution of relations extracted by GPT-4 remained further from the test set distribution (one-sided Mann-Whitney U test, p-value = 9.25e-05, Wasserstein distance = 0.356) compared to BioBERT, which exhibited a closer match to the test set distribution (Wasserstein distance = 0.180).

Considering the reported superior performance of advanced closed-source models like GPT-4 in other generative tasks in biomedical domain and reports suggesting that their training data likely includes extensive biomedical literature\cite{nori2023capabilities,kwon2024publishers}, we hypothesize that its shortcomings in biomedical text mining partially stem from insufficient understanding of dataset-specific requirements, rather than from limitations in understanding biomedical documents themselves. To validate this, we conducted a case study focusing on the failures of LLMs in these tasks (Table 2). 

\setlength{\LTleft}{-1cm}  
\setlength{\LTright}{0cm}  

\begin{longtable}{p{4.5cm} p{13.5cm}}
\caption{Major Failure Patterns and Case Studies}
\label{tab:failures} \\
\toprule
\textbf{Failure Pattern} & \textbf{Explanations and Examples} \\
\midrule
\endfirsthead

\multicolumn{2}{l}{\textit{(Continued from previous page)}} \\
\toprule
\textbf{Failure Pattern} & \textbf{Explanations and Examples} \\
\midrule
\endhead

\bottomrule
\multicolumn{2}{l}{\textit{(Continued on next page)}} 
\endfoot

\bottomrule
\endlastfoot

\textbf{1. Semantically Plausible but Inconsistent with Annotation Guidelines} 
& 
Predictions appear reasonable but fail to comply with the annotation guideline details, leading to incorrect labels. Examples include following:

\vspace{1cm}
\textbf{(a) Incorrect Annotation Hierarchy.} \\ 
& 
\begin{itemize}
\setlength{\itemsep}{10pt}
    \item In the NCBI-Disease dataset, both mentions of specific \textit{diseases} and \textit{disease classes} (families of specific diseases) should be annotated. However, GPT-4 using prompt same as prior research sometimes neglected disease class entities. For example: 
    \begin{quote}
    \textit{``The RB1 gene mutation was investigated in a child with ectopic intracranial retinoblastoma using DNA obtained from both the pineal and retinal tumours of the patient.''}(PMID: 9400934\cite{onadim1997rb1})
    \end{quote}
    the LLM properly annotated all other \textit{disease} mentions but failed to label ``\textit{pineal and retinal tumours}'' as a disease class.
\end{itemize}
\\[1ex]
& 
\textbf{(b) Incorrect Annotation Category.} \\
& 
\begin{itemize}
    \item In BioRED (RE) dataset, relationships must be classified into categories like ``positive correlation,'' ``negative correlation,'' or ``association.'' However, the LLM sometimes misinterpreted the direction of effect. For example:
    \begin{quote}
    \textit{``\ldots intravenous lidocaine infusion or a high dose of mexiletine \ldots controlled the ventricular tachycardia.''}(PMID: 15485686\cite{chang2004novel})
    \end{quote}
    The LLM labeled the pair \textit{(mexiletine, ventricular tachycardia)} as ``positive correlation.'' While it is true that mexiletine has a beneficial therapeutic effect on this disease, the guidelines specify that when a drug is used to treat a disease or reduces its severity, it should be labeled as ``negative correlation''. 
\end{itemize}
\\[1ex]
& 
\textbf{(c) Incorrect Boundary or Granularity.} \\
& 
\begin{itemize}
    \item Certain datasets (e.g., BC5CDR-Disease) require separate mention annotations for a full disease name and any parenthetical abbreviations. However, the LLM occasionally merges both into a single entity mention. For instance, from ``\textit{Parkinson's disease (PD)},'' two separate mentions (``\textit{Parkinson's disease}'' and ``\textit{PD}'') are required, but the LLM collapses them into a single mention.
\end{itemize}
\\[1ex]
& 
\textbf{(d) Neglecting Special Exceptions.} \\
& 
\begin{itemize}
\setlength{\itemsep}{10pt}
    \item According to the guideline of NCBI-Disease, words like \textit{``disease''}, \textit{``syndrome''}, or \textit{``deficiency''} alone should not be annotated. However, in the following:
    \begin{quote}
    \textit{``Low levels of beta hexosaminidase A in healthy individuals with apparent deficiency of this enzyme.''}(PMID: 941901\cite{navon1976low})
    \end{quote}
    the LLM incorrectly annotated ``\textit{deficiency}'' as a disease despite the guideline explicitly prohibiting that. 
\end{itemize} \\

\midrule
\textbf{2. Errors due to Inadequate Semantic Comprehension} 
&
Some errors are purely due to reasoning mistakes by the LLM and are independent of annotation guidelines. Examples are shown in Appendix Table S1. Interestingly, for the case in Table S1.2, the LLM produced correct answers in a free-text style when given a simpler prompt, instead of one requiring detailed structured outputs. This suggests that forcing structured outputs may disrupt the LLM's internal reasoning. This observation motivated our proposed technique in Section \ref{sec:twostep}.

\\[1ex]

\midrule

\textbf{3. Parsing Errors} 
& 


Parsing errors occur when an LLM’s output does not follow the required structured format, causing parsing failures. While models like LLaMA2-13B were reported to frequently exhibit such issues \cite{chen2023large,touvron2023llama}, we did not encounter them outside the development stage. This can be attributed to that frontier LLMs benefit from extensive training focused on generating structured outputs, as well as that we implemented a robust output parser capable of capturing edge cases. 

\\

\end{longtable}

\section{Methodology}
To address the challenges identified in Section \ref{sec:LLMchallenge}, we applied a series of prompt engineering strategy for biomedical text mining tasks. This strategy includes general techniques like dynamic few-shot prompting, as well as some targeted methods like two-step inference for maintaining both reasoning processes alongside structured output, automatic prompt optimization to account for distributional differences between LLM outputs and ground truth labels, and a workflow for dynamically extracting annotation guidelines to provide detailed instructions to LLMs. 

The experiments mainly utilized the GPT-4o model (2024-08-06), which was one of the most advanced and widely used models. Additionally, in the later stages of this study, several LLMs supporting test-time compute were released, which are not constrained by missing reasoning steps due to formatting limitations. Therefore, we also evaluated OpenAI o1-mini which is a representative model in this category (\ref{o1mini}).

To ensure that the observed performance improvements were not solely due to advancements in the underlying LLM itself, we also conducted experiments on the LitCovid and BC5CDR-Chemical datasets using historical model consistent with previous studies (see Appendix 1.2). Moreover, we employed model distillation on the HoC and BC5CDR-RE datasets by annotating recent articles with LLMs to construct augmented datasets, which were then used to fine-tune BERT-based models.

Below, we describe these components in detail.

\subsection{Prompt Engineering}
\subsubsection{Dynamic Few-shot Prompting}
\textit{In-Context Learning} \textit{In-Context Learning} (ICL) is a foundational capability of LLMs, enabling them to perform new tasks given instructions or a few examples in the input. The choice of examples significantly impacts model performance \cite{nori2023capabilities}. We adopted a semantic-based dynamic few-shot prompting approach, which is a widely adopted technique across diverse tasks.

\begin{figure}[H] 
\centering
\hspace*{-20px}
\includegraphics[width=0.6\textwidth]{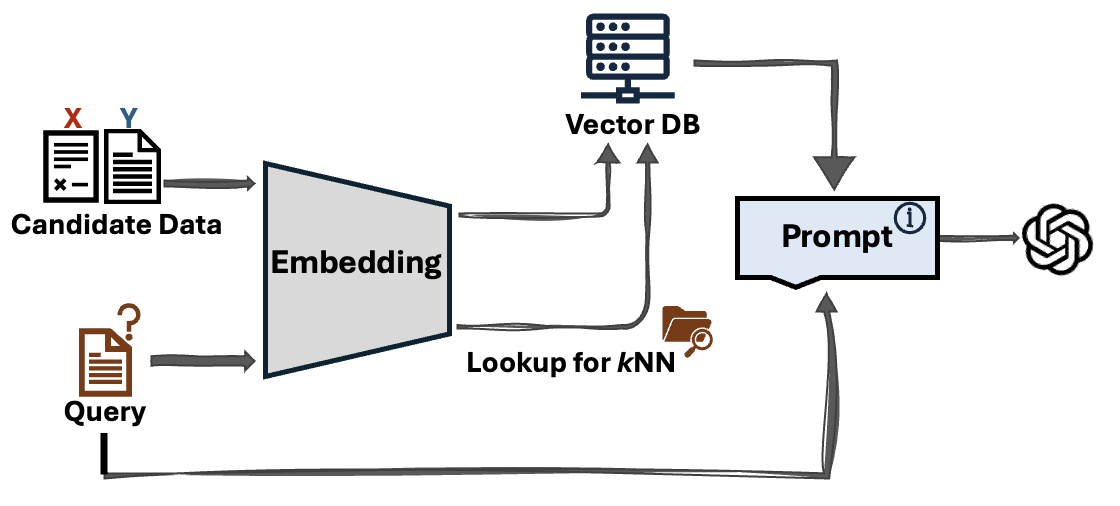}
\caption{\textbf{Workflow of the Dynamic Few-shot Prompting}}
\label{fig:fewshot} 
\end{figure}

As showned in Figure~\ref{fig:fewshot}, we embedded each document in the dataset and stored these them in a vector database. In the inference phase, the $K$ nearest documents in the embedding space, based on cosine similarity, were retrieved as examples along with their expected outputs. We used OpenAI’s \texttt{text-embedding-3-small} model for embeddings and Chroma as the vector database.

\subsubsection{Strategies for Structured Output}
For biomedical multi-label text classification, we adopt JSON as the standard output format. JSON is widely recognized as a structured data format for output generation in LLM and is explicitly supported by APIs such as OpenAI's. In our implementation, each key in the JSON structure corresponds to a label, and its associated value indicates whether the label is selected.

However, JSON is not well-suited for NER and RE tasks. For NER, the conventional approach involves the BIO format\cite{ramshaw1995text}, which uses the labels B (Beginning), I (Inside), and O (Outside) to denote whether a token—typically a word or subword—marks the start of an entity, is inside an entity span, or lies outside any entity span. This approach aligns with the pre-LLM paradigm for NER, where the task is framed as sequence labeling. Prior LLM evaluation using BIO format for NER showed a very limited performance\cite{jahan2024comprehensive}. LLMs, by design, excel at generating coherent text rather than making predictions for individual tokens. To address this, S. Wang et al. developed a method that uses delimiters to enclose entities\cite{wang2023gpt}. Building on this insight and considering that contemporary LLM training datasets often include a significant amount of HTML-formatted text, we adopted an HTML tagging scheme for NER. In our approach, entities are marked using tags such as: \texttt{<Type = "{\(\{entity\ type\}\)}"> {\(\{entity\ mention\}\)} </Type>}

For RE tasks, which typically involve two entities and their relationship, we represent relationships using tuples. In the BC5CDR-RE task, where only one relationship type (Chemical-Disease Interaction, CDI) is involved, we use a format like: \texttt{({\(entity_1\)}, {\(entity_2\)})}. In contrast, for BioRED, which includes multiple relationship types, we use the format of triples: \texttt{( {\(entity_1\)}, {\(entity_2\)}, {\(relation\_type\)} )}.

\subsubsection{Two-step Inference to Maintain Both Reasoning and Structured Outputs}\label{sec:twostep}
Chain-of-Thought (CoT) prompting enhances reasoning by explicitly incorporating intermediate steps before providing final answers \cite{wei2022chain}. Frontier LLM are often trained on datasets that include reasoning processes, enabling them to autonomously decompose problems and engage in step-by-step reasoning even without explicit prompting\cite{liu2024datasets}. However, in biomedical text mining tasks, models are typically required to extract structured results directly from query text in bulk, rather than generating unstructured, difficult-to-parse output. Such omissions can degrade model performance (e.g. Appendix Table S1.b). Quantitative analyses by Z. Tam et al. further corroborate similar findings\cite{tam2024let}.

\begin{figure}[H] 
\centering
\hspace*{-20px}
\includegraphics[width=0.5\textwidth]{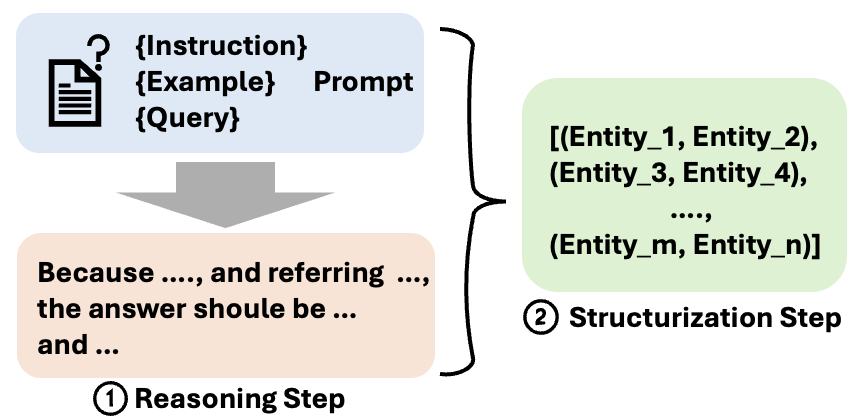}
\caption{\textbf{Workflow of the Two-step Inference to Maintain Both Reasoning Step and Structured Outputs}}
\label{fig:twostep} 
\end{figure}

Some recently released models that support test-time computation do not inherently suffer from this issue. However, for widely used models released before the test-time computation era, which offer faster inference (such as GPT-4o), structured constraints often force the model to bypass intermediate reasoning processes. To resolve this, we introduce a two-step approach that allows LLMs to first generate complete reasoning processes, and subsequently extract structured answers from them (Figure \ref{fig:twostep}). For multi-label biomedical text classification tasks requiring JSON outputs, we inserted an "intermediate steps" schema before the final output in OpenAI’s structured output API\cite{openaistructure}. For NER and RE tasks which demand alternative output formats, and for the experiment in Appendix 1.2, we implemented this functionality independently.

\subsubsection{Automatic Prompt Optimization}
In Section \ref{sec:LLMchallenge}, our analysis reveals a discrepancy between the distribution of labels predicted by LLM and the ground truth labels, and the inability of capturing such implicit features may contribute to their suboptimal performance. To address this, we aim to optimize prompts by narrowing the gap between predicted and correct label distributions.

We adopt and modify R. Pryzant's pseudo-“gradient descent” approach which was inspired by gradient descent but the target of optimizatation is the prompt instead of the model parameters\cite{pryzant2023automatic}. This method identifies distributional gaps on small batches of data and constructs a natural-language “gradient” critiquing the current prompt. By editing the prompt semantically opposite to this gradient, we iteratively refine prompts that optimize the dataset-level metric (Macro-F1). Multiple candidate prompts are generated in each iteration, using beam search to select top candidates for subsequent refinement rounds. Given its high computational cost, our experiments were limited to the LitCovid and BC5CDR-Chemical datasets. Detailed descriptions are provided in Appendix 1.3. due to space constraints.



\begin{figure}[H] 
\centering
\hspace*{-20px}
\includegraphics[width=0.4\textwidth]{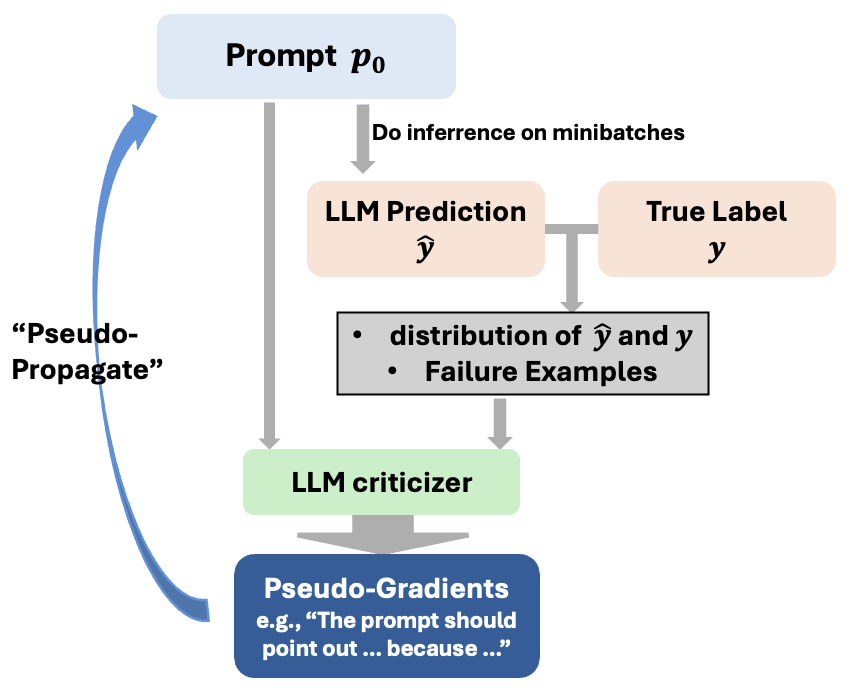}
\caption{\textbf{Automatic Prompt Optimization Using Pseudo "Gradient Descent"} }
\label{fig:automatic} 
\end{figure}

\subsection{Instruction Retrieval from Annotation Guideline}

Our analysis in Section \ref{sec:LLMchallenge} reveals that the limitations of LLMs in biomedical text mining arise primarily from their insufficient comprehension of  detailed, dataset-specific annotation requirements. Initially, we attempted to integrate complete annotation guidelines into the prompts, but yielded negligible improvements, possibly due to the excessive length of the guidelines hindering the LLM’s effective utilization of information within long contexts.


Thus, we explored dynamically extracting relevant entries from the annotation guidelines for each query case. A common approach for such tasks is Retrieval-Augmented Generation (RAG), which integrates retrieval mechanisms with external knowledge bases to provide LLMs with accurate contextual information\cite{gao2023retrieval, fan2024survey}. While RAG techniques are already implemented in advanced commercial generative models, we could not find a readily available framework suitable for our specific needs. Therefore, we developed a custom rule-based pipeline tailored to these requirements.

\begin{figure}[H] 
\centering
\hspace*{-20px}
\includegraphics[width=0.75\textwidth]{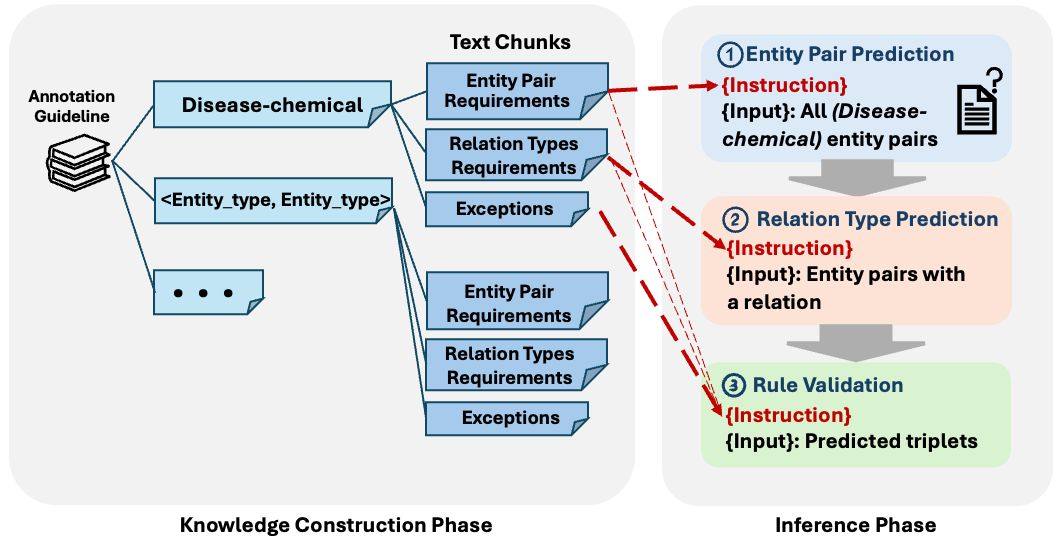}
\caption{\textbf{Instruction Retrival Workflow in BioRED (RE) Dataset}}
\label{fig:RAG} 
\end{figure}

For example, in the BioRED (RE) dataset, the annotation guidelines span 16 pages, with around 10 pages dedicated to relationship extraction. Within these, the “relation pairs” section accounts for about 4 pages, and the “relation types” section covers 6 pages. These requirements are further divided into seven subsections based on entity types (e.g., “Disease-chemical,” “Disease-variant”). Specific rules for edge cases are grouped in a separate “Others” subsection. To facilitate retrieval, we reorganized these guidelines into text chunks (Figure~\ref{fig:RAG}, Knowledge Construction Phase).


During inference, processing each test query involves three steps (Figure~\ref{fig:RAG}, Inference Phase). Alos, instead of querying each candidate entity pairs individually, all candidate pairs from two specific entity types are queried simultaneously to reduce inference costs.
\begin{enumerate} 
\item \textbf{Relation(Entity Pair) Prediction}: Predicts if relationships exist between entity pairs using the relevant \textit{Relation Pair Requirements}.

\item \textbf{Relation Type Classification}: Classifies the relation types of identified pairs using the relevant \textit{Relation Type Requirements}. 

\item \textbf{Rule Validation}: Checks whether predicted pairs and types require special handling and conform to guideline constraints, ensuring accuracy. \end{enumerate}

For other datasets accompanied by annotation guidelines, we adopted a similar workflow. For the BC5CDR-Chemical/Disease (NER) and NCBI-Disease (NER) datasets, we first provided the LLM with simplified requirements on annotating entity mentions, manually summarized from the annotation guidelines, instructing the model to generate preliminary annotations. Subsequently, we incorporated the complete annotation guidelines including examples and edge cases into the prompt, guiding the LLM for detailed rule validation. For the BC5CDR relation extraction (RE) dataset, since its annotation guideline is concise (less than half a page), we directly integrated the guideline into the prompt, enabling the LLM to produce results with a single query.

\subsection{Model Distillation}

To explore the potential of LLMs in replacing human annotators in the widely adopted “manual annotation–model training–inference” workflow in real-world applications, we conducted model distillation from GPT-4o to a BERT-based model using data augmentation and weak supervision, which are fundamental strategies in LLM distillation.\cite{xu2024survey}.

For the HoC dataset, we used the original search keywords to retrieve PubMed articles published after April 2024 via NCBI’s E-utilities API\cite{sayers2009utilities, PubMed}. We annotated these articles with GPT-4o using two-step inference, then filtered out articles without positive labels to construct an LLM-annotated corpus of 3,000 articles. This synthetic dataset was then used to fine-tune a BioLinkBERT-Large model.

Similarly, for the BC5CDR-RE dataset, following the same data collection procedure as in the original study, we retrive all related articles published recently. Using the titles and abstracts of these papers, we first performed named entity recognition to extract entity lists, followed by relation extraction to identify chemical-disease interactions (CDI) utilizing GPT-4o with two-step inference + instruction retrieval from the annotation guideline. We created an augmented corpus of 2,361 articles containing 5,930 CDI records, which was then used to fine-tune BioLinkBERT-Large. Negative samples were generated by randomly pairing Chemical-Disease entities without CDI relationships, maintaining a 2:1 negative-to-positive ratio.

The timeframe after April 2024 was selected to minimize overlap between retrieved articles and the GPT-4o’s training data. Furthermore, this approach mimics a real-world scenario where novel annotation needs arise, requiring the labeling of newly published articles. 

\section{Results and Discussion}

\subsection{Evaluation of GPT-4o Performance With Proposed Methods}
Tables \ref{tab:cls}, \ref{tab:ner}, and \ref{tab:re} summarize GPT-4o’s performance across with targeted prompt engineering. Additionally, five out of six failure cases listed in Table \ref{tab:failures} and Appendix Table 1 (PMIDs: 24717468, 15485686, 941901, 15036754, 22836123) were successfully addressed using the “Two-step Inference + Instruction Retrieval from Annotation Guideline” approach. The remaining case (PMID: 9400934) was resolved using the “Dynamic 3-shot + Instruction Retrieval from Annotation Guideline” method.

\begin{table}[H]
\centering
\caption{Results on Multi-label Text Classification Tasks}
\label{tab:cls}
\renewcommand{\arraystretch}{1.5}
\begin{tabular}{p{5cm} ccc ccc}
\hline
 & \multicolumn{3}{c}{LitCovid} 
 & \multicolumn{3}{c}{HoC} \\
\hline
 & \textbf{Precision} & \textbf{Recall} & \textbf{F1} 
 & \textbf{Precision} & \textbf{Recall} & \textbf{F1} \\
\hline
Basic Instruction  
 & 0.663 & 0.892 & 0.744
 & 0.705 & 0.896 & 0.797 \\
Dynamic 3-shot     
 & 0.739 & 0.860 & 0.798
 & 0.720 & 0.857 & 0.786 \\
\makecell[l]{Dynamic 3-shot \\+ Automatic Prompt Optimization}
 & 0.815 & 0.839 & 0.822
 & -     & -     & -     \\
Two-step Inference
 & 0.831 & 0.909 & \textbf{0.873}
 & 0.827 & 0.865 & 0.841 \\
\hline
Baseline (SOTA Foundation model)
 & 0.870 & 0.849 & 0.861
 & 0.887 & 0.864 & \textbf{0.873} \\
\hline
\end{tabular}
\end{table}

\begin{table}[H]
\caption{Results on Named Entity Recoginization Tasks}
\label{tab:ner}
\hspace*{-1.8cm}
\renewcommand{\arraystretch}{1.5}
\begin{tabular}{p{7cm} ccc ccc ccc}
\hline
 & \multicolumn{3}{c}{BC5CDR-Chemical} 
 & \multicolumn{3}{c}{BC5CDR-Disease} 
 & \multicolumn{3}{c}{NCBI-Disease} \\
\hline
 & \textbf{Precision} & \textbf{Recall} & \textbf{F1} 
 & \textbf{Precision} & \textbf{Recall} & \textbf{F1} 
 & \textbf{Precision} & \textbf{Recall} & \textbf{F1} \\
\hline
Basic Instruction  
 & 0.814 & 0.739 & 0.775
 & 0.629 & 0.441 & 0.519
 & 0.710 & 0.645 & 0.676 \\
Dynamic 3-shot     
 & 0.828 & 0.789 & 0.809
 & 0.711 & 0.572 & 0.634
 & 0.700 & 0.690 & 0.695 \\
\makecell[l]{Dynamic 3-shot \\+ Automatic Prompt Optimization}
 & 0.829 & 0.821 & 0.825
 & -     & -     & -
 & -     & -     & -     \\
\makecell[l]{Dynamic 3-shot \\+ Instruction Retrieval from Annotation Guideline}
 & 0.849 & 0.841 & 0.845
 & 0.789 & 0.567 & 0.659
 & 0.803 & 0.734 & 0.767 \\
Two-step Inference
 & 0.829 & 0.872 & 0.850
 & 0.761 & 0.622 & 0.684
 & 0.771 & 0.744 & 0.757 \\
\makecell[l]{Two-step Inference \\+ Instruction Retrieval from Annotation Guideline}
 & 0.920 & 0.885 & 0.902
 & 0.733 & 0.789 & 0.760
 & 0.843 & 0.780 & 0.810 \\
\hline
Baseline (SOTA Foundation model)
 & 0.939 & 0.942 & \textbf{0.940}
 & 0.856 & 0.876 & \textbf{0.866}
 & 0.882 & 0.912 & \textbf{0.897} \\
\hline
\end{tabular}
\end{table}

\begin{table}[H]
\centering
\caption{Results on Relation Extraction Tasks}
\label{tab:re}
\renewcommand{\arraystretch}{1.5}
\begin{tabular}{p{6cm} ccc ccc}
\hline
 & \multicolumn{3}{c}{BC5CDR (RE)} 
 & \multicolumn{3}{c}{BioRED} \\
\hline
 & \textbf{Precision} & \textbf{Recall} & \textbf{F1} 
 & \textbf{Precision} & \textbf{Recall} & \textbf{F1} \\
\hline
Basic Instruction  
 & 0.639 & 0.679 & 0.618
 & 0.323 & 0.259 & 0.274 \\
Dynamic 3-shot     
 & 0.662 & 0.646 & 0.654
 & 0.392 & 0.292 & 0.318 \\
\makecell[l]{Dynamic 3-shot \\+ Instruction Retrieval from Annotation Guideline}
 & 0.632 & 0.656 & 0.647
 & 0.615 & 0.417 & 0.497 \\
Two-step Inference
 & 0.732 & 0.801 & 0.741
 & 0.474 & 0.354 & 0.406 \\
\makecell[l]{Two-step Inference \\+ Instruction Retrieval from Annotation Guideline}
 & 0.761 & 0.833 & \textbf{0.781}
 & 0.613 & 0.528 & 0.567 \\
\hline
Baseline  
 & \makecell[l]{\textbf{BioBERT}: 0.693\\ \;\;\textbf{BioGPT}: 0.495}
 & \makecell[l]{0.615\\0.432}
 & \makecell[l]{0.650\\0.462}
 & 0.571 & 0.605 & \textbf{0.589} \\
\hline
\end{tabular}
\end{table}

The \textbf{Dynamic 3-shot} approach outperformed the basic instruction prompts (i.e., prompts similar to those used in prior studies) across all datasets, except for HoC. This can be attributed to that more contextually relevant examples provided the LLM with additional background knowledge and annotation style about similar queries. However, this method still indirectly relies on manually annotated data. Conversely, in the \textbf{Two-step Inference} (Reasoning Step \& Structurization Step) approach, few-shot prompting was not utilized due to the impracticality of pre-writing reasoning process for every instance in the training set. Nevertheless, the Two-step Inference method achieved significant performance improvements across all tested datasets. This highlights again that the commonly used structuralization for LLMs in biomedical text mining tasks can severely constrain the reasoning capabilities that frontier LLMs acquired during their training phase.

We also experimented with \textbf{automatic prompt optimization}, iteratively refining prompts by using natural language to describe the distributional differences between predicted and correct labels (resembling a "pseudo-gradient descent" approach). However, the performance gains were modest compared to the huge computational cost, likely because the discrepancy patterns between predicted and correct labels varied across test instances. As a result, a short natural language "gradient" struggled to generalize across the test set. Analyzing the prompts generated during optimization revealed that many included generic but dataset-irrelevant instructions (e.g., rewriting prompts in Markdown, adding overly detailed background introductions) or directly retelled the pseudo-gradient information, such as "\textit{the predicted label distribution on the test set was …, while the actual distribution was… Use this information to ..."}" Such instructions were clearly unusable by the LLM during inference. Nevertheless, some iterations produced potentially helpful guidance, such as "\textit{Be mindful of over-predicting topics such as …. These should only be selected if …}". Given the computational cost and limited performance improvement, we cannot recommend this approach for production scenario in biomedical text mining. However, it could serve as a source of inspiration when manually curating prompts.

Instruction Retrieval from Annotation Guidelines generally improved performance, except when combined with Dynamic 3-shot on the BC5CDR (RE) dataset. Its impact was particularly notable for datasets like BioRED, which involve complex entity-relation annotations, or when paired with the Two-step Inference approach. Unfortunately, we could not apply this method to datasets such as LitCovid and HoC due to the absence of detailed annotation guidelines. 

Overall, implementing systematic enhancements targeted for biomedical text mining enabled GPT-4o to outperform baseline models on two datasets with minimal manual annotation dependency and achieve competitive performance on others. This challenges the prevailing notion in prior studies that LLMs underperform in discriminative tasks within the biomedical domain.

\subsection{Evaluation of Reasoning Model (o1-mini)}\label{o1mini}

Table 6 presents our experimental results on the reasoning model (OpenAI o1-mini). Except for the fact that the model’s built-in test-time computation replaces the need for Two-Step Inference, all prompting settings remain the same as those used with GPT-4o. Across all datasets, o1-mini performs slightly worse than GPT-4o enhanced with proposed approach. We speculate that larger models, such as o1 or Deepseek-R1\cite{guo2025deepseek}, which are capable of deeper reasoning, could achieve higher performance than o1-mini. However, due to computational costs and API providers’ constraints on processing time, we were unable to include these models in our experiments. 

\begin{table}[ht]
\centering
\begin{threeparttable}
\caption{Comparison of o1-mini and GPT-4o on various datasets.}
\label{tab:comparison}
\renewcommand{\arraystretch}{1.2}
\begin{tabular}{l ccc ccc}
\hline
 & \multicolumn{3}{c}{\textbf{o1-mini} $^*$} & \multicolumn{3}{c}{\textbf{GPT-4o}} \\
\hline
\textbf{Dataset} & \textbf{Precision} & \textbf{Recall} & \textbf{F1} 
                & \textbf{Precision} & \textbf{Recall} & \textbf{F1} \\
\hline
LitCovid        & 0.824 & 0.895 & 0.837 & 0.831 & 0.909 & 0.873 \\
HoC             & 0.770 & 0.849 & 0.807 & 0.827 & 0.865 & 0.841 \\
BC5CDR-Chemical & 0.767 & 0.828 & 0.797 & 0.920 & 0.885 & 0.902 \\
BC5CDR-Disease  & 0.681 & 0.683 & 0.708 & 0.733 & 0.789 & 0.760 \\
NCBI-Disease    & 0.777 & 0.763 & 0.770 & 0.843 & 0.780 & 0.810 \\
BC5CDR (RE)     & 0.685 & 0.722 & 0.703 & 0.761 & 0.833 & 0.871 \\
BioRED          & 0.465 & 0.448 & 0.456 & 0.613 & 0.528 & 0.567 \\
\hline
\end{tabular}
\begin{tablenotes}\footnotesize
\item[$^*$] \textit{Since the reasoning model does not require "Two-Step Inference" to retain the thought process, we utilized Dynamic 3-shot prompting for multi-label text classification tasks (LitCovid, HoC), and "Instruction Retrieval from Annotation Guideline" prompting for NER tasks (BC5CDR-Chemical/Disease, NCBI-Disease) and RE tasks (BC5CDR RE, BioRED).}
\end{tablenotes}
\end{threeparttable}
\end{table}

\subsection{Result of Model Distillation}

Following original procedures, we retrieved recent papers and constructed synthetic training datasets for HoC and BC5CDR-RE annotated by GPT-4o.

BioLinkBERT-Large fine-tuned on these synthetic datasets achieved a Macro-F1 of 0.817 on HoC, slightly below its performance on the smaller, expert-annotated official training set (Table \ref{tab:distillation}). For BC5CDR-RE, the distilled BioLinkBERT model achieved a Macro-F1 of 0.646, closely approaching the performance of 0.650 obtained by fine-tuning on the official training set. Notably, it outperformed the fine-tuned BioGPT, which achieved a Macro-F1 score of 0.462 and was reported as the SOTA method on this dataset.

\begin{table}[H]
\centering
\caption{Results about Model Distillation on HoC and BC5CDR-RE Datasets}
\label{tab:distillation}
\renewcommand{\arraystretch}{1.3}
\begin{tabular}{p{5cm} ccc ccc}
\hline
 & \multicolumn{3}{c}{HoC} 
 & \multicolumn{3}{c}{BC5CDR-RE} \\
\hline
 & \textbf{Precision} & \textbf{Recall} & \textbf{F1} 
 & \textbf{Precision} & \textbf{Recall} & \textbf{F1} \\
\hline
\makecell[l]{Distillated Model\\(BioLinkBERT-Large Fine-tuned\\on Augmented Dataset)} & 0.797 & 0.835 & 0.817
 & 0.625 & 0.668 & 0.646 \\
\hline
\makecell[l]{Baseline\\(BioLinkBERT-Large Fine-tuned\\on Trainset)}
 & 0.887 & 0.864 & \textbf{0.873}
 & 0.693 & 0.615 & \textbf{0.650} \\
\hline
\end{tabular}
\end{table}

\begin{figure}[H] 
\centering
\hspace*{-20px}
\includegraphics[width=0.45\textwidth]{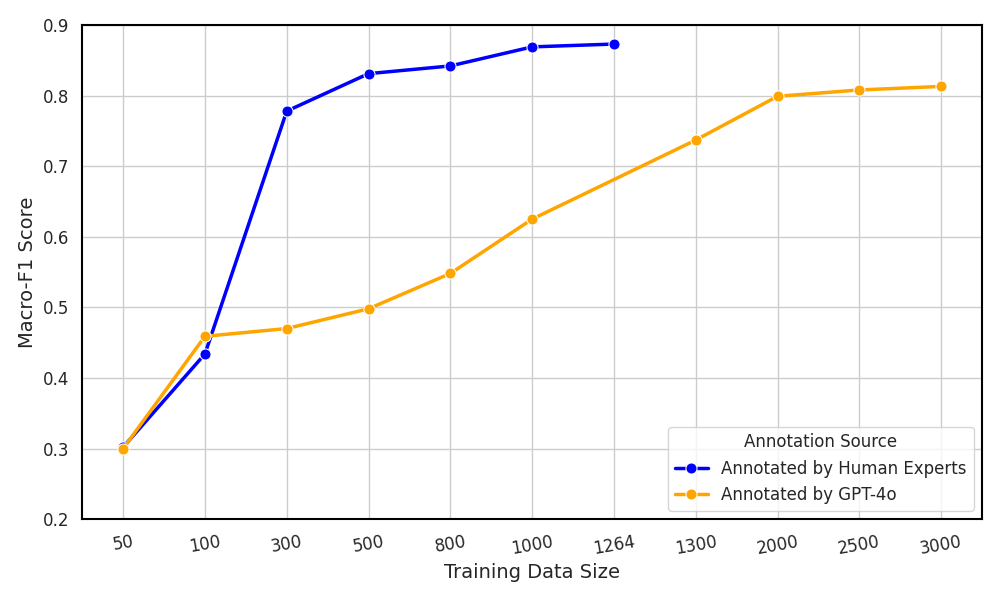}
\caption{\textbf{Relationship between Training Data Size and Performance on HoC Dataset}}
\label{fig:dis} 
\end{figure}

Moreover, we investigated the relationship between the size of supervised data and the performance of BERT-based model. As shown in Figure \ref{fig:dis}, the performance of BioLinkBERT improves rapidly with an initial increase in supervised data. However, the benefit of additional data diminishes beyond a certain level, and the performance approaches a plateau. Manually annotated datasets enable the BERT model to achieve higher performance with relatively smaller data volumes. Nonetheless, leveraging large-scale data labeled through stacked prompts from LLMs can yield performance levels close to those obtained with manual annotation, while significantly incurring lower human resource costs.

Based on these findings, we propose that in production scenarios requiring continuous inference of a large amount of texts but with less stringent performance demands—such as document classification to assist with reading—LLMs guided by targeted prompt engineering have the opportunity to replace human annotators in the widely adopted “manual annotation–model training–inference” workflow. This approach offers a cost-effective alternative while maintaining acceptable levels of performance for many practical applications.

\section{Conclusions and Future Work}
Our study analyzed factors contributing to LLMs' suboptimal performance in biomedical text mining using conventional methods. We subsequently developed targeted prompting strategies to address these challenges. While core ideas of these prompt engineering techniques and information retrieval frameworks were not originally ours, we customized them to better suit the specific needs of those tasks. Results demonstrated that frontier LLMs, like GPT-4o, could challenge the prior notion of their unsuitability for biomedical discriminative tasks. 

Our results indicate that combining two-step inference and instruction retrieval from annotation guidelines can greatly enhance LLM performance. 
However, the current method relies heavily on predefined rules for guideline segmentation and information retrieval, requiring substantial preparatory and coding efforts. Future improvements could involve developing an LLM annotator agent that automatically queries the necessary guidelines and examples, or integrating a trainable retriever like in Li et al.’s research\cite{li2024biomedrag} to dynamically extract guidelines, instead of the rule-based approach. Additionally, we did not fine-tune any open-source LLM due to computational limitation, but the proposed strategies and model fine-tuning approaches are complementary, and their combination may potentially yield even better performance.

In production applications, for tasks requiring highly accurate data annotation—such as integrating information extracted from literature data into a biological database—LLMs, like existing machine learning methods, cannot guarantee performance at the level of human experts. Even though we anticipate that future iterations and upgrades of foundational models (including reasoning models) will further enhance LLM performance in biomedical text mining, reaching human expert-level accuracy remains a significant challenge. Nonetheless, the inference process of LLMs can serve as a reference for human annotators, potentially reducing their workload.

Furthermore, for applications that can tolerate some degree of annotation errors (e.g., improving literature readibility), LLMs, despite performance variations, still demonstrate practical predictive capabilities. One of the key advantages of LLMs is their ability to "cold-start" when new annotation needs arise—such as new annotation objects, types, or texts—without relying on supervised data constructed by human annotators. However, the inference cost of LLMs can become a concern when annotating large volumes of text, such as labeling all papers in the PubMed database, making model distillation a practical alternative.

Additionally, since our evaluations were conducted on commonly used biomedical text mining datasets, we cannot entirely rule out the risk of data contamination (Appendix 1.4). Considering this, we recommend continuous monitoring and management of LLM performance when deploying this methodology in production environments.

\section{Data Availability}
The data and prompt can be found at the following repository: 
\url{https://github.com/ekkkkki/LLM-Replace-Annotators-in-Biomedical-Text-Mining}.

\bibliography{main}

\section{Appendix}
\subsection{Additional Information about Analysis of LLM Failure}\label{appendix:detail1}

\renewcommand{\thefigure}{S\arabic{figure}}
\setcounter{figure}{0}

\renewcommand{\thetable}{S\arabic{table}}
\setcounter{table}{0}

\begin{figure}[H] 
\centering
\includegraphics[width=0.7\textwidth]{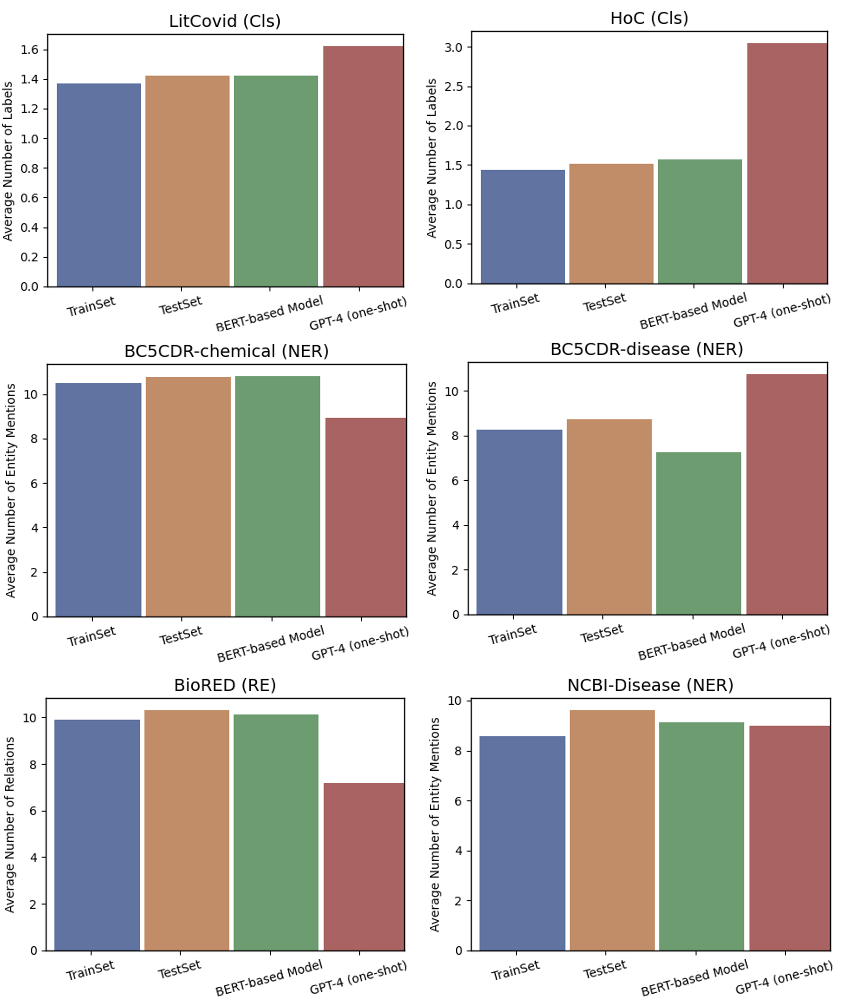}
\caption{\textbf{Average Number of Labels, Entity Mentions, or Relations of Trainset, Testset, BERT Output, and LLM Output Across Various Datasets}
}
\label{fig:s1}
\end{figure}

\begin{figure}[H] 
\centering
\includegraphics[width=0.8\textwidth]{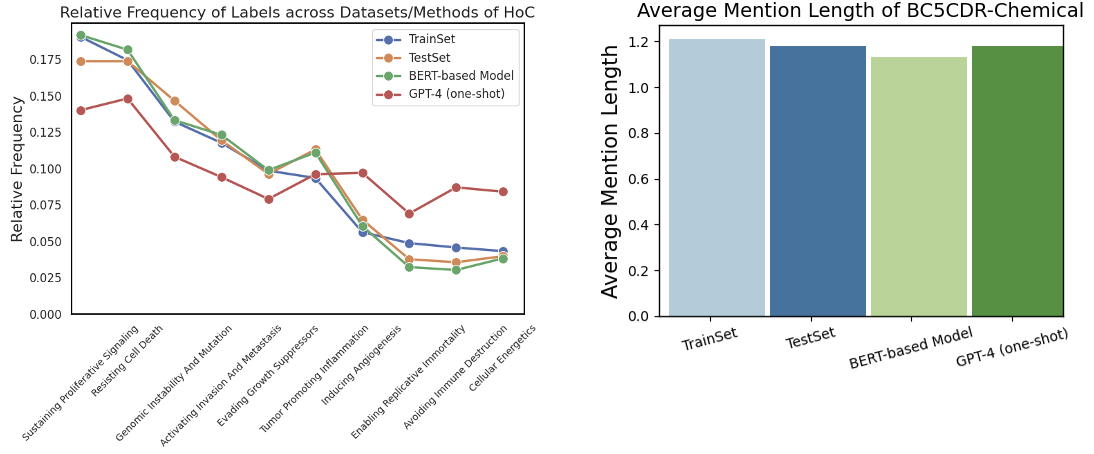}
\caption{\textbf{Relative Frequency of Labels on HoC and Average Mention Length of BC5CDR-Chemical}
}
\label{fig:s2}
\end{figure}

\begin{longtable}{p{4.5cm} p{13.5cm}}
\caption{Extra Examples for Failure Pattern Studies}
\label{tab:failures} \\
\toprule
\textbf{Failure Pattern} & \textbf{Examples} \\



\bottomrule

\textbf{1. Semantically Plausible but Inconsistent with Annotation Guidelines} 
& 
\textbf{An example of Incorrect Annotation Hierarchy.} 
\begin{itemize}
\setlength{\itemsep}{10pt}
    \item In the BioRED (RE) dataset, only specific relationships (e.g., disease--chemical) among five defined entity types should be annotated. Nevertheless, the LLM labeled ``\textit{dexmedetomidine}'' and ``\textit{patient}'' as having an ``\textit{Association}'' relationship in: 
    \begin{quote}
    \textit{``A comparison of severe hemodynamic disturbances between dexmedetomidine and propofol for sedation in neurocritical care patients.''}(PMID: 24717468\cite{erdman2014comparison})
    \end{quote}
    Where according to the guidelines, ``\textit{patient}'' should be marked as a ``\textit{Species}'' entity, which should not be linked in any relationship with a chemical entity in this dataset context.
\end{itemize}
\\[1ex]
& 
\textbf{An Example of Neglecting Special Exceptions.} \\
& 
\begin{itemize}
\setlength{\itemsep}{10pt}
    \item For the following sentence in  BioRED (RE) dataset: 
    \begin{quote}
    \textit{``The acute toxicity of OPs is the result of their irreversible binding with AChEs in \ldots''}(PMID: 15036754\cite{tuovinen2004organophosphate})
    \end{quote}
    LLM annotated ``\textit{OP}'' and ``\textit{toxicity}'' as having a ``negative correlation,'' even though the annotation guideline requires not to annotate the chemicals with the disease concept-“toxicity” directly, but to annotate the disease due to the “toxicity” and “drug-induced” instead.
\end{itemize} \\

\midrule
\textbf{2. Errors due to Inadequate Semantic Comprehension} 
&
In the following article snippet from BC5CDR (RE) dataset:
\vspace{0.2cm}
\begin{quote}
\textit{``Late-onset scleroderma renal crisis induced by tacrolimus and prednisolone: a case report. Scleroderma renal crisis (SRC) is a rare complication of systemic sclerosis (SSc) but can be severe enough to require temporary or permanent renal replacement therapy. Moderate to high dose corticosteroid use is recognized as a major risk factor for SRC. Furthermore, there have been reports of thrombotic microangiopathy precipitated by cyclosporine in patients with SSc.''}(PMID:22836123\cite{nunokawa2014late})
\end{quote}
\vspace{0.2cm}
The correct chemical--disease interactions (CDI) that should be extracted include:
\textit{Cyclosporine--microangiopathy, Corticosteroid--SRC/SSc, tacrolimus--SRC/SSc}. 
However, the LLM omitted \textit{(tacrolimus, SRC/SSc)} and misidentified the disease associated with ``\textit{cyclosporine}'' as SRC/SSc instead of \textit{microangiopathy}.
\vspace{0.2cm}

In this example, with a simpler prompt setting (e.g., ``\textit{retrieve all the CID pairs from the text}''), the LLM successfully predicted the correct pairs in a more free-text style.

\\[1ex]

\midrule

\end{longtable}


\subsection{Experiments using Historical Models}\label{sec:historical}
To ensure that the observed performance improvements of LLM in our results are not solely attributable to the model upgrading itself, we conducted evaluations using the same model and API versionas those employed in Qingyu Chen et al.'s research \cite{chen2023extensive} (Table \ref{tab:s1}, \ref{tab:s2}). In their experiments, GPT-4 0613 achieved an F1 score of 0.608 on the LitCovid dataset, which is lower than the results we replicated. On the BC5CDR-chemical dataset, it reached a F1 score of 0.833, comparable to our experimental outcomes.

Furthermore, when employed with targeted prompt engineering, the performance of GPT-4 0613 demonstrated significant improvements that the gap with GPT-4o is less than 2\% on the BC5CDR-chemical dataset. On the LitCovid dataset, the performance disparity was more pronounced, which may be attributed to the limitations of GPT-4 0613 in customizing schemas within its structured output API to incorporate intermediate reasoning steps. As a result, we had to implement a manual two-step inference mechanism, which might have influenced the outcomes.

\begin{table}[H]
\centering
\caption{Results on LitCovid with GPT-4o and Historical Model}
\label{tab:s1}
\renewcommand{\arraystretch}{1.5}
\begin{tabular}{p{5cm} ccc ccc}
\hline
 & \multicolumn{3}{c}{GPT-4o 0806} 
 & \multicolumn{3}{c}{GPT-4 0613} \\
\hline
 & \textbf{Precision} & \textbf{Recall} & \textbf{F1} 
 & \textbf{Precision} & \textbf{Recall} & \textbf{F1} \\
\hline
Basic Instruction  
 & 0.663 & 0.892 & 0.744
 & 0.778 & 0.649 & 0.707 \\
Dynamic 3-shot     
 & 0.739 & 0.860 & 0.798
 & 0.661 & 0.894 & 0.760 \\
Two-step Inference
 & 0.831 & 0.909 & 0.873
 & 0.701 & 0.881 & 0.809 \\
\hline
\end{tabular}
\end{table}

\begin{table}[H]
\centering
\caption{Results on BC5CDR-Chemical with GPT-4o and Historical Model}
\label{tab:s2}
\renewcommand{\arraystretch}{1.5}
\begin{tabular}{p{7cm} ccc ccc}
\hline
 & \multicolumn{3}{c}{GPT-4o 0806} 
 & \multicolumn{3}{c}{GPT-4 0613} \\
\hline
 & \textbf{Precision} & \textbf{Recall} & \textbf{F1} 
 & \textbf{Precision} & \textbf{Recall} & \textbf{F1} \\
\hline
Basic Instruction  
 & 0.814 & 0.739 & 0.775
 & 0.855 & 0.699	& 0.770 \\
Dynamic 3-shot     
 & 0.828 & 0.789 & 0.809
 & 0.837 & 0.786 & 0.810 \\
Two-step Inference
 & 0.829 & 0.872 & 0.850
 & 0.828 & 0.829 & 0.828 \\
\makecell[l]{Two-step Inference \\+ Instruction Retrieval from Annotation Guideline}
 & 0.920 & 0.885 & 0.902
 & 0.902 & 0.856 & 0.881 \\
\hline
\end{tabular}
\end{table}

\subsection{Automatic Prompt Optimization}
Numerous automatic prompt engineering methods have been developed\cite{sahoo2024systematic, chen2023unleashing}, and we used R. Pryzant's pseudo-“gradient descent” approach which was inspired by gradient descent but the target of optimizatation is the prompt instead of the model parameters\cite{pryzant2023automatic}. In this method, we identify distributional gap between LLM predictions and ground truth distributions on small batches of data, along with errors in the LLM's output, to construct a natural language "gradient" that critiques the current prompt. By editing the prompt in the semantic direction opposite to this gradient, we aim to propagate the gradient into the prompt and identify a prompt that optimizes the desired metric (Macro-F1 score) at the dataset level. During the prompt refinement process, multiple candidate prompts are generated, and beam search is employed as the search algorithm. In each iteration, the top \textit{k} prompts with the highest scores are retained as candidates for the next round, while lower-scoring prompts are discarded.

However, this method is computationally expensive, primarily due to the need for iterative batch inferences. To achieve reliable estimates of distribution differences, we set the batch size to 100. Additionally, implementing this approach for different tasks requires significant coding effort. As a result, our experiments were limited to the LitCovid and BC5CDR-Chemical dataset. For each iteration, we generated three pseudo-gradients along with corresponding candidate prompts during the "propagation" step. In the beam search, the top three candidates were retained in each iteration, and a total of 10 iterations were performed.

\subsection{Data Contamination}\label{sec:contami}
The pre-training datasets used by the frontier commercial LLMs remain largely undisclosed. However, it is widely believed, based on information from various news reports, that these datasets likely include biological research articles\cite{pooley2024large,gibney2024has,kwon2024publishers}. Our examination of the data collection processes in openly available LLM datasets suggests that the possibility of data contamination is relatively low for datasets stored on FTP servers in formats like JSONL or PubTator (e.g., the BioRED dataset)\cite{openlm2023openllama,awesomehealthcare}. However, there is also a possibility that the correct labels could have been inadvertently leaked through web elements or other means\cite{li2024task}, as exemplified by the LitCovid website, which aggregates categorical labels for articles.

\end{document}